\documentclass[letterpaper, 10 pt, conference]{ieeeconf}  

\IEEEoverridecommandlockouts                              

\usepackage{graphics} 
\usepackage{amsmath} 
\usepackage{amssymb}  
\graphicspath{ {./images/} }
\usepackage{amsmath,amsfonts, physics}
\usepackage{algorithmic}
\usepackage{array}
\usepackage{textcomp}
\usepackage{stfloats}
\usepackage{graphicx}
\usepackage{hyperref}
\usepackage{multirow}
\usepackage{amssymb}
\usepackage{float}
\title{\LARGE \bf
Micromanipulation in Surgery: Autonomous Needle Insertion Inside the Eye for Targeted Drug Delivery
}

\author{Ji Woong Kim$^{1}$, Peiyao Zhang$^{1}$, Peter Gehlbach$^{2}$, Iulian Iordachita$^{1}$, Marin Kobilarov$^{1}$ 
\thanks{$^{1}$Johns Hopkins University, Laboratory for Computational Sensing and Robotics (LCSR), Baltimore, Maryland 21218, USA}%
\thanks{$^{2}$Wilmer Eye Institute, Johns Hopkins University, Baltimore, Maryland 21218, USA}%
}

\begin{document}

\maketitle
\thispagestyle{empty}
\pagestyle{empty}

\begin{abstract}

We consider a micromanipulation problem in eye surgery, specifically retinal vein cannulation (RVC). RVC involves inserting a microneedle into a retinal vein for the purpose of targeted drug delivery. The procedure requires accurately guiding a needle to a target vein and inserting it while avoiding damage to the surrounding tissues. RVC can be considered similar to the “reach” or “push” task studied in robotics manipulation, but with additional constraints related to precision and safety while interacting with soft tissues. Prior works have mainly focused developing robotic hardware and sensors to enhance the surgeons' accuracy, leaving the automation of RVC largely unexplored. In this paper, we present the first autonomous strategy for RVC while relying on a minimal setup: a robotic arm, a needle, and monocular images. Our system exclusively relies on monocular vision to achieve precise navigation, gentle placement on the target vein, and safe insertion without causing tissue damage. Throughout the procedure, we employ machine learning for perception and to identify key surgical events such as needle-vein contact and vein punctures. Detecting these events guides our task and motion planning framework, which generates safe trajectories using model predictive control to complete the procedure. We validate our system through 24 successful autonomous trials on 4 cadaveric pig eyes. We show that our system can navigate to target veins within 22$\mu m$ of XY accuracy and under 35 seconds, and consistently puncture the target vein without causing tissue damage. Preliminary comparison to a human demonstrates the superior accuracy and reliability of our system.
\end{abstract}

\begin{figure}
        \centering
        \includegraphics[width = \columnwidth]{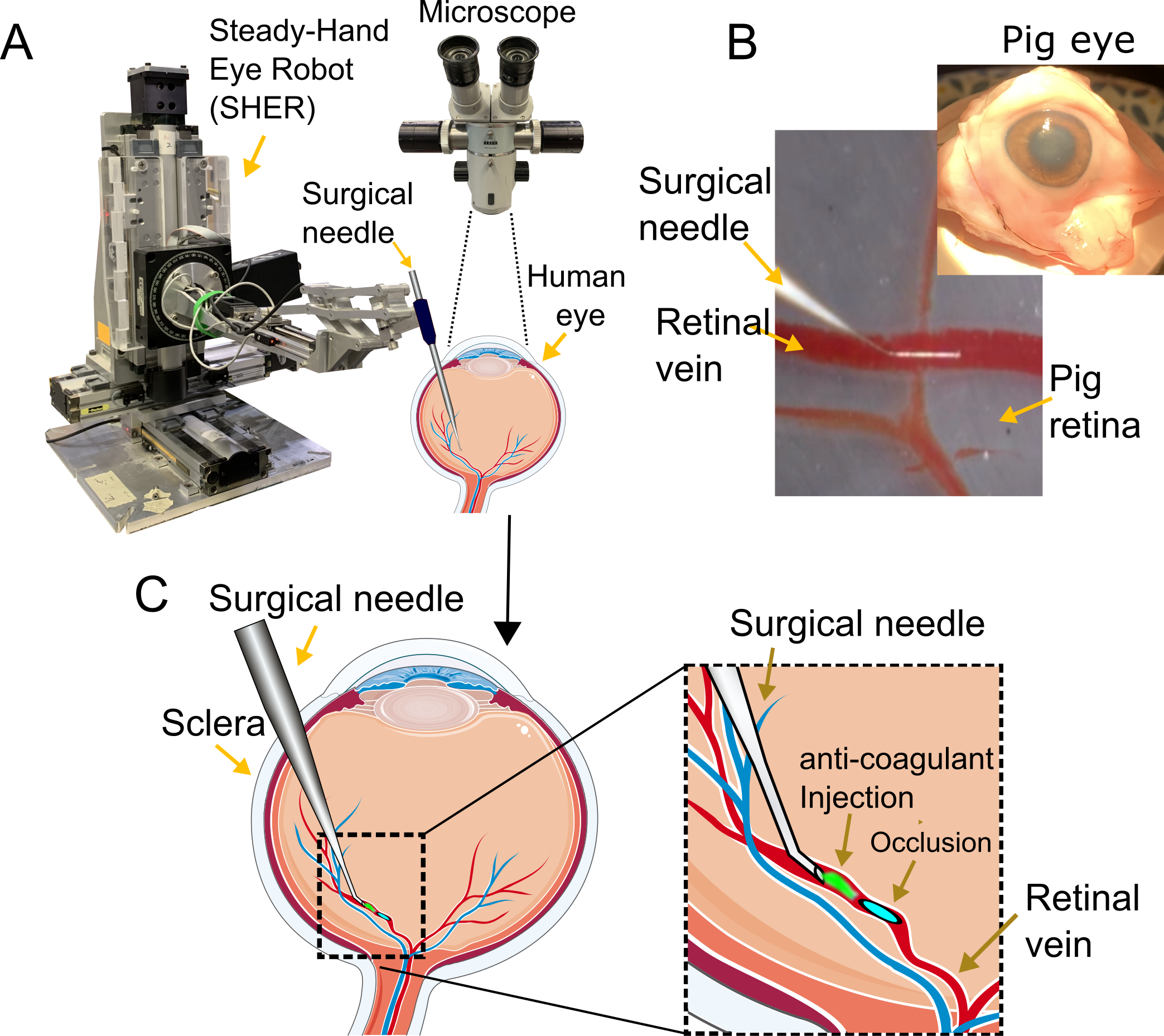}
        \vspace{-5mm}
        \caption{(A) Illustration of retinal vein cannulation; a surgical needle is inserted into the eye via a point incision on the sclera. A microscope is suspended over the patient's eye to visualize the retina. (B) Cadaveric pig eye used in the experiments (C) Illustration of retinal vein cannulation procedure; a microneedle is penetrated into a small blood vessel (typically ranging from 60 - 120 microns in diameter) to inject anti-coagulants and dissolve an occlusion downstream of the vein.}
        \label{fig:intro}
\end{figure}

\section{INTRODUCTION}
We consider a micromanipulation problem in retinal surgery, specifically retinal vein cannulation (RVC). RVC involves inserting a microneedle into a retinal vein for the purpose of targeted drug delivery. RVC begins with the surgeon introducing the needle through a small  point incision on the sclera of the eye, while visualizing the retina using a top-down microscope. The surgeon then precisely navigates the needle inside the eye, which is filled with a jelly-like substance known as the vitreous, and gently land the needle-tip at the desired blood vessel. This step is equivalent to a “reach” task encountered in the robotic manipulation literature. Then, the needle is inserted along its axis to cause vein-puncture, after which the drug is infused into the vein for up to several minutes.

RVC is an important problem as it may serve as a potential cure to retinal vein occlusion (RVO), a blinding disease caused by a blockage in the retinal vein and which affects 16.4 million adults world-wide \cite{rvo_prevalence}. By performing RVC, a small dosage of anti-coagulant is delivered into the bloodstream to chemically dissolve the blockage. The result is restoration of blood flow into the eye and restored vision for the patient. However, current existing treatments only provide preventative measures, and the only definite potential cure is performing RVC. However, RVC poses difficult challenges, including: (1) navigation to the target vein within tens-of-micrometers of accuracy, (2) depth estimation challenge due to the top-down view of the surgery, (3) constrained surgical workspace of approximately 1 inch sphere, and (4) the non-regenerative nature of the retina. Given the risks and the navigation accuracy that is nearly impossible to achieve by hand, RVC is rarely practiced clinically. Prior works have focused on developing robotic platforms to reduce the surgeons’ hand-tremor or developing sensors that warn surgeons of adverse events \cite{robotic_surgery_review} \cite{fbg_force_sensing},  \cite{OCT_integrated_tool} \cite{puncture_detection_impedance}, and some exploration of automation \cite{eye_surgery_imitation_learning} \cite{peiyao_eye_surgery} \cite{peiyao_pig_eye_paper}, \cite{mach2022oct}, \cite{autonomous_navigation_retina}, \cite{wei2022region}, \cite{kim2023ioct}. However, automation of this procedure largely remains unexplored.

In this paper, we propose a first autonomous strategy for this procedure using a minimal setup: a surgical robot, a needle, and monocular images. Note that we solve this problem while solely relying on visual cues obtained from monocular image feedback, without estimating depth. At a high-level, we solve the navigation problem by visual-servoing the needle to the target vein until a contact cue between the needle-tip and the vein is \emph{visually} observed. Specifically, upon contact with the vein, the needle-tip deflects and  changes appearance. We detect this contact event using a template matching algorithm. This algorithm effectively enables placement of the needle-tip on the target vein without causing damage. Then, the needle is inserted along its axis. During insertion, a recurrent convolutional neural network (CNN) is used to detect a vein-puncture event, so that the needle can be stopped immediately upon detecting puncture. This is to avoid double-puncturing the vein and potentially damaging the underlying retina. 
\begin{figure*}[]
        \centering
        \includegraphics[width = \textwidth]{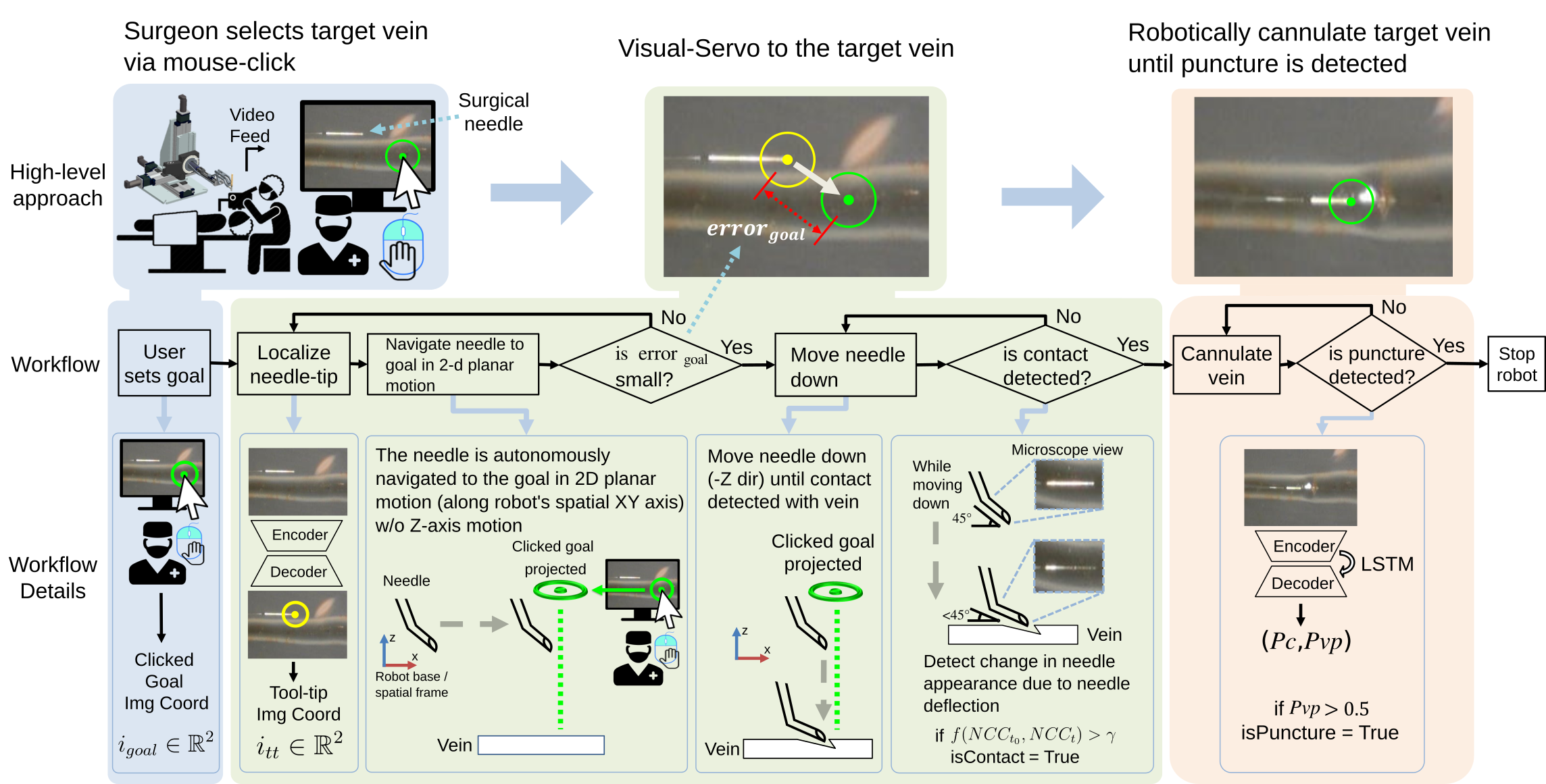}
        \vspace{-5mm}
        \caption{Autonomous retinal vein cannulation workflow: initially, the surgeon selects the desired insertion goal via a mouse-click in a monocular visual feed. Then, the needle-tip is navigated to the clicked goal via planar motion, such that the needle-tip is aligned with the clicked goal in the microscope view. Then, the needle is simply lowered towards the vein via motion along the Z-axis of the robot. When the needle-tip makes contact with the target vein, the needle deflects and changes appearance, which is used as a cue to stop the robot. Then, the needle is inserted along its axis, and during this step, a recurrent CNN is used to detect a puncture event to stop the robot.}
        \label{fig:workflow}
\end{figure*}

The entire workflow is implemented using a state-machine framework. Specifically, we identify the main steps of RVC and encode them as states in the state-machine framework. Then, various vision-based algorithms are used to extract the current surgical state information and detect surgical events such as needle-vein contact and vein-puncture. This information triggers transitions between surgical states and initiates appropriate surgical actions to be autonomously executed. Throughout the autonomous steps, model predictive control (MPC) is used to generate optimal trajectories while satisfying safety constraints. One such safety constraint is the remote-center-of-motion (RCM) at the entry point into the eye. This constraint ensures that extraneous rotation is minimized at the entry point (i.e. the needle can only slide and tilt about the entry point) to avoid causing damage to the scleral tissue.


To validate our system, we demonstrate 24 successful autonomous RVC trials on 3 cadaveric pig eyes, and show preliminary comparison results compared to a human operator. Our contribution is as follows:

\begin{enumerate}
    \item A first autonomous strategy for RVC, while relying solely on monocular visual guidance. The system is experimentally validated through 24 successful autonomous trials on 3 pig eyes, achieving on average 24$\mu m$ XY navigation accuracy and a total duration of 35 seconds. 
    \item A strategy for visually detecting needle-tissue interactions such as physical contact and vein-punctures, which was not previously studied in literature.
    \item Preliminary comparison to a human operator in robot-assisted and free-hand mode. Our autonomous system achieves superior navigation accuracy and safer needle insertion trajectory, with much lower risk for damaging the vein.
\end{enumerate}

\begin{figure*}[]
        \centering
        \includegraphics[width = \textwidth]{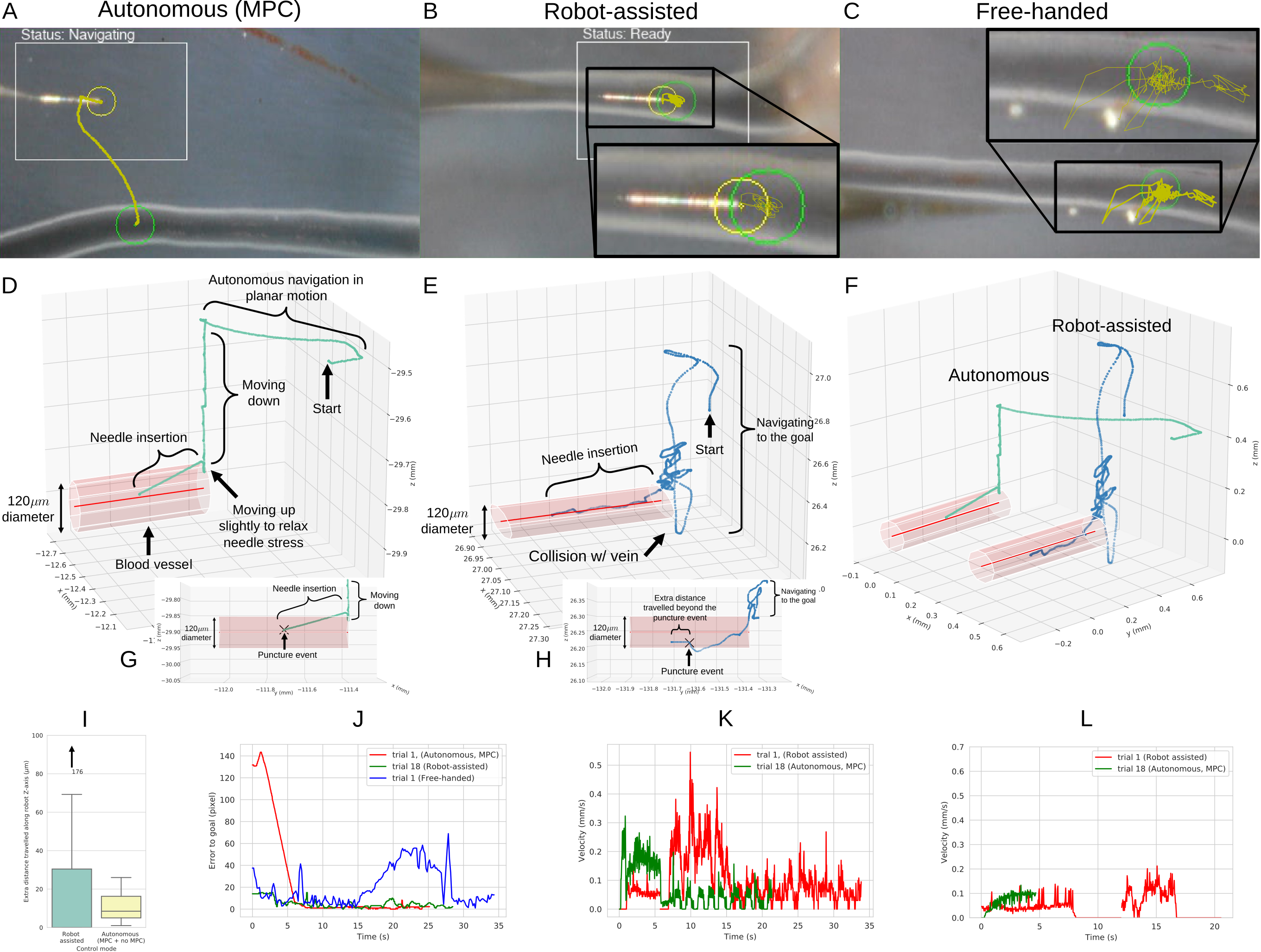}
        \caption{Navigation trajectory is shown for (A) autonomous, (B) robot-assisted, and (C) free-hand modes. (D, E) The trajectories of the autonomous mode and the free-hand mode is shown in 3D. (F) Their trajectories are compared side-by-side. (G, H) A close-up of the insertion trajectory is shown from a side view (I) Extra distance travelled along the Z-axis after the puncture is compared for robot-assisted and autonomous trials. (J) The error to the final clicked goal is compared between robot-assisted and autonomous mode. (K) The velocity of the needle navigation trajectory is compared between robot-assisted and autonomous mode. (L) Similar comparison is made during needle insertion}
        \label{fig:results}
\end{figure*}

\subsection{Method}
Our experimental setup consists of a surgical robot, a surgical needle with a tip diameter of 15$\mu m$, and a top-down microscope generating observations of the surgical workspace (Fig. \ref{fig:intro}). Cadaveric pig eyes were used for testing. The eyes were cut in half and the veins were injected with air to simulate vascular pressure of live veins. 

The high-level workflow is demonstrated in Fig. \ref{fig:workflow}. At a high-level, our system consists of a perception and planner module. The perception module detects relevant information necessary for automating the task, such as detecting the surgical tool-tip position, contact events, and vein-puncture events. The planner uses this information to determine the current state of the surgery, identify the required action to perform, and outputs an optimal trajectory using MPC for task completion. In this paper, we focus our discussion on the needle-vein contact and vein-puncture parts. 

\subsubsection{Detecting needle-vein contact}

Needle-vein contact is detected using the flexible property of the needle. Specifically, as the needle is lowered and makes contact with the target vein, the needle bends about its elbow (the needle has a 45 degree bend), and this leads to a visual change in the microscope image as shown in Fig. \ref{fig:workflow}. We detect this change using a template matching algorithm with an added layer of logic. Specifically, a small template image of the needle is acquired, which is then compared to the needle appearance in the subsequent time steps. For the similarity metric, we use the normalized cross-correlation (NCC) metric, which is essentially a normalized dot product between the two images. After the needle significantly changes appearance, the NCC score drastically decreases, and we detect this decreasing trend to detect the contact event. Specifically, if the percent change in the NCC score is larger than a chosen threshold value, then contact event is detected. More specifically, the percent change in maximum NCC score compared to its initial score is given by:

\begin{equation}
    f(NCC_{t_0}, NCC_t) = \frac{max(NCC_{t_0}) - max(NCC_{t})}{max(NCC_{t_0})}
\end{equation}
\normalsize

where $NCC_{t_0}$ is the NCC heat map at the start of the needle lowering motion and $NCC_t$ is the NCC heat map calculated in subsequent time steps. If the percent change of maximum NCC score is greater than a chosen threshold gain, then the algorithm detects a contact event. In other words,

\begin{equation}
    isContact = 
\begin{cases}
    True ,& \text{if }  f(NCC_{t_0}, NCC_t) \geq \gamma \\
    False,              & \text{otherwise}
\end{cases}
\end{equation}

The threshold gain $\gamma$ was chosen by observing 14 representative vein-contact events as examples. Specifically, we manually chose a value such that contact events could be detected with a small delay in these examples.

 \subsubsection{Detecting venipuncture}
 Next, we discuss how vein-puncture events are detected. The puncture event occurs with sudden release of potential energy as the needle quickly enters the lumen. This event is visible to the human eye and usually occurs with a sudden popping motion. Our initial attempt was to design a logic-based approach to detecting this puncture event. However, there were too many variations at which punctures occurred, sometimes even occluding the needle-tip. Therefore, instead of designing a hand-crafted algorithm to detect puncture, we collected 250 puncture examples to train a recurrent neural network to detect punctures. The input to the network is the current surgical image and the output is a binary vector denoting the probability of whether puncture has occurred or not. The network uses a Resnet-18 backbone \cite{resnet}. Throughout the data collection process, we noticed that some punctures occurred very subtly and less dramatically than others. Thus, to make the puncture more easily detectable by the network, we subsampled the input video from 30Hz to 7Hz. 

 To improve the network's performance, we added a decoder network on top of the encoded features to predict the input image, mimicking an auto-encoder network \cite{unet}. Such use of auxiliary loss added an additional gradient signal that improved the network's performance on the main task of detecting punctures. At the time of inference, however, the decoder was ignored and thus it did not add any additional computational overhead during inference. We express the loss function as 

\begin{equation}
\begin{split}
    L((y, I), (\hat{y}, \hat{I})) =  & -\frac{1}{N} \sum_{n=1}^{N} \{ y_i \cdot log(\hat{y_i}) \\
    & + (1-y_i) \cdot log(1-\hat{y_i}) + (I - \hat{I})^2 \}
    \end{split}
\end{equation}
where $y$ is the binary indicator for the true class, $\hat{y}$ is the predicted probability of puncture, $I$ is the input image, and $\hat{I}$ is the image predicted by the decoder part of the network. We define $p_c = \hat{y}$ and $p_{vp} = 1 - \hat{y}$ as the probability of contact and venipuncture respectively. Note that the puncture detection module is activated after detecting a needle-vein contact event. Thus, $p_c$ is not used to detect needle-vein contact events. It simply denotes the probability of the current surgical state being the surgical state preceding the event of puncture.

\subsection{Results}

To demonstrate the efficacy of our system, we performed a total of 24 needle insertion trials on 3 cadaveric pig eyes. All trials were successful, though the system can be made susceptible to failure due to perception errors. Based on these experiments we report that on average navigation error to the clicked goal is 24 microns along XY directions. This error is calculated by comparing the needle-tip position to the clicked goal pixel in the microscope image, and converting this pixel error to microns using a conversion factor of 136.33pixels/mm. We also report the delay in detecting contact and punctures. Specifically, when needle-vein contact or vein-puncture occurs, they are not detected immediately but after some temporal delay. Because the robot is still in motion during this window of delay, we measure the amount of extra distance travelled during this period. Our experiments show that on average, the contact detection delay is 40 microns, and puncture delay along Z-axis is 26 microns on average (Z-axis is pointed towards the second wall of the vein, thus it is the most safety-critical direction), which are acceptable given the 60 - 120 micron diameter of retinal veins. 

Comparison to a robot-assisted, and free-hand trials are shown in Fig. \ref{fig:results}. In robot-assisted mode, the surgeon controls the surgical needle by directly operating at the robot handle and using a gain pedal to control the velocity of the motion. It can be seen that the autonomous trajectory is efficient, however, the human operator struggles to stabilize at the desired needle insertion point and fails to maintain a consistent angle during needle insertion in robot-assisted mode. If the needle-tip is not maintained at an optimal angle, this may require the needle greater travel distance to puncture the vein, increasing the risk of double-puncturing the vein. In free-hand mode, it was extremely difficult to stabilize the needle at the desired insertion goal. While it was possible to successfully insert the needle into the vein in a few occasions, inserting the needle at the pixel goal location was nearly impossible due to the uncontrollable hand tremor.

\section{CONCLUSION}

In this work, we demonstrate an autonomous system for RVC. Our system requires minimal setup and guidance, requiring monocular images as input and goal specified by the user via clicking. We showed that our system could accurately navigate the surgical needle onto the target vein within required margins of safety, successfully perform needle insertion, and detect a vein-puncture event in a timely manner to avoid double-puncturing the vein. We demonstrated the consistency of our system and its ability to generalize  across various pig eye anatomy. 

In this work, we did not consider the patient's eye motion. During the surgery, the patient has motion related to breathing, which causes cyclic oscillation of the retinal tissue. This may not be a critical limitation since in recent robot-assisted human trials, the needle was held static inside the vein without motion compensation \cite{first_human_retinal_surgery}.

Future work will consider extending this method with OCT to ensure greater safety and accuracy during the procedure, and with greater robustness to moving conditions.

\addtolength{\textheight}{-12cm}   

\bibliography{bib}  




\end{document}